\documentclass[letterpaper]{article}
\usepackage{aaai}
\usepackage{times}
\usepackage{helvet}
\usepackage{courier}
\usepackage{graphicx}
\usepackage{threeparttable}
\usepackage{ntheorem}
\usepackage{color}
\usepackage{colortbl}
\usepackage{multirow}
\usepackage{amsmath}
\usepackage{amsfonts}
\usepackage{comment}
\usepackage{bm}                                            
\usepackage{etoolbox}                                      
\usepackage{url}
\usepackage{nth}
\usepackage{cite}
\usepackage{natbib}
\usepackage{balance}
\usepackage[bookmarks=false]{hyperref}                     %
\usepackage{courier}                                       
\usepackage{listings}                                      
\usepackage[]{algpseudocode}                               
\algrenewcommand\textproc{\texttt}
\makeatletter\let\c@float@type\relax\makeatother
\makeatletter\let\float@addtolists\relax\makeatother
\usepackage{algorithm}
\hypersetup{
    colorlinks = true,
    citecolor  = blue,
    linkcolor  = blue,
	urlcolor   = blue,
}

\newcommand{\tabincell}[2]{                                
	\begin{tabular}{@{}#1@{}}#2\end{tabular}
}

\makeatletter
\newcommand{\thickhline}{%
	\noalign {\ifnum 0=`}\fi \hrule height 1pt
	\futurelet \reserved@a \@xhline
}

\theoremstyle{plain}
\theoremheaderfont{\normalfont\bfseries}\theorembodyfont{\slshape}
\theoremsymbol{\ensuremath{\diamondsuit}} \theoremseparator{~}
\newtheorem{mytheorem}{Theorem}

\newenvironment{myproof}{\emph{Proof.}}{\hfill$\square$}

\RequirePackage[normalem]{ulem} 
\RequirePackage{color}\definecolor{RED}{rgb}{1,0,0}\definecolor{BLUE}{rgb}{0,0,1} 

\graphicspath{{./figs/}}

\begin{document}

\title{Robust Matrix Regression}
\author{{\bf $\text{Hang Zhang}^1$, $\text{Fengyuan Zhu}^1$, $\text{Shixin Li}^2$ } \\
$~^1$ Department of Computer Science and Engineering, The Chinese University of Hong Kong \\
$~^2$ College of Electronic and Information Engineering, Sichuan University \\
hzhang@cse.cuhk.edu.hk, fyzhu@cse.cuhk.edu.hk, leept416@gmail.com}

\maketitle

\begin{abstract}
	\begin{quote}
		Modern technologies are producing datasets with complex intrinsic structures, and they can be naturally represented as matrices instead of vectors.
		To preserve the latent data structures during processing, modern regression approaches incorporate the low-rank property to the model, and achieve satisfactory performance for certain applications.
		These approaches all assume that both predictors and labels for each pair of data within the training set are accurate. 
		However, in real world applications, it is common to see the training data contaminated by noises, which can affect the robustness of these matrix regression methods. 
		In this paper, we address this issue by introducing a novel robust matrix regression method. 
		We also derive efficient proximal algorithms for model training.
		To evaluate the performance of our methods, we apply it on real world applications with comparative studies. 
		Our method achieves the state-of-the-art performance, which shows the effectiveness and the practical value of our method. 
	\end{quote}
\end{abstract}
\section{Introduction}

Classical regression methods, such as ridge regression~\citep{ML1970_hoerl1970_ridge} and lasso~\citep{ML1996_tibshirani_regression} are basically designed for data in vector form. 
However, with the development of modern technology, it is common to meet datasets with sample unit not in vector form but instead in matrix form. 
Examples include the two-dimensional digital images, with quantized values of different colors at certain rows and columns of pixels; and electroencephalography~(EEG) data with voltage fluctuations at multiple channels over a period of time. 
When using traditional regression methods to process these data, we have to reshape them into vectors, which may destroy the latent topological structural information, such as the correlation between different channels for EEG data~\citep{ML2014_zhou_regularized}, and the spatial relation within an image~\citep{ML2007_wolf_modeling}. 

To tackle this issue, several methods have been proposed to perform regression on data in matrix form directly.
One such model is the regularized matrix regression~(R-GLM)~\citep{ML2014_zhou_regularized}. Given a dataset $ \{ \mathbf{X}_{i}, y_i \}_{i=1}^{N} $, where $ N $ is the sample size, $ \mathbf{X}_i \in \mathbb{R}^{p\times q} $ denotes the $ i $th data matrix as predictor, and $ y_i \in \mathbb{R} $ is the corresponding output, the R-GLM model aims to learn a function $ f: \mathbb{R}^{p\times q} \rightarrow \mathbb{R}  $ to identify the output given a newly observed data matrix with
\begin{equation}
\label{eqn:RGLM}
y_i = \text{tr}(\mathbf{W}^T\mathbf{X}_i) + b + \epsilon,
\end{equation}
where $ \text{tr}(\cdot) $ represents the trace of a matrix, $ W $ is the regression matrix with low-rank property to preserve structural information of each data matrix, and $b$ denotes the offset. $ \epsilon $ is zero-mean Gaussian noise to model the small uncertainty of the output. With this setting, the R-GLM has achieved satisfying results in several applications.
However, there still exist certain issues that should be further addressed.
Firstly, R-GLM uses the Gaussian noise for model fitting, and take all deviations of predicted values from labels into account. This setting can be reasonable in certain cases, but may not make sense for particular applications. 
As an example, consider the problem of head pose estimation~\citep{ML2001_sherrah_fusion}, where for each data pair, the predictor is a two dimensional digital image for the head of a person, while the output denotes the angle of his head. 
Because the real angle of head cannot be measured precisely, there should exist certain deviations of provided labels from the real ones empirically. In this case, the regression model should be able to tolerant such small deviations instead of taking them all into account. 
Another important issue is that, the predictors are also assumed to be noise free, which can be irrational in certain applications. Practically, it is common to see signals corrupted by noise, such as image signals with occlusion, specular reflections or noise~\citep{ML2016_huang_robust}, and financial data with noise~\citep{ML1998_magdon_financial}. Thus, it is important for a regression model to be tolerant of noise on predictors and labels to enhance its robustness empirically.  
 
In this paper, we introduce two novel matrix regression methods to tackle the above mentioned issues. 
We first propose a ``Robust Matrix Regression''~(RMR) to tackle the noisy label problem, by introducing hinge loss to model the uncertainty of regression labels. In this way, our method only considers error larger than a pre-specified value, and can tolerate error around each labeled output within a small range. This approach is also favored for other advantages in certain scenarios. 
As an example, in applications like financial time-series prediction, it is common to require not to lose more than $ \epsilon $ money when dealing with data like exchange rates, and this issue can be well addressed with our setting. Even though the hinge loss error has been used in the support vector regression model~\citep{ML2004_smola_tutorial}, it is an algorithm based on vector-form data, which can ruin the latent structure for matrix regression problem. 
We then propose efficient ADMM method to solve the optimization problem iteratively. 

To further enhance the robustness of RMR with noisy predictors, we propose a generalized RMR~(G-RMR) by decomposing each data matrix as latent clean signal plus sparse outliers. For model training, we also derive a proximal algorithm to estimate both the regression matrix and latent clean signals iteratively. 
To evaluate the performance of our methods, we conduct extensive experiments on both approaches with comparison of state-of-the-art ones. Our methods achieve superior performance consistently, which shows their efficiency in real world problems.

\textbf{Notations}: We present the scalar values with lower case letters~(e.g., $x$); vectors by bold lower case letters~(e.g., $ \mathbf{x} $); and matrix by bold upper case letters~(e.g., $ \mathbf{X} $). For a matrix $ \mathbf{X} $, its $ (i,j) $-entity is represented as $ \mathbf{X}_{i,j} $. $ \text{tr}(\cdot) $ denotes the trace of a matrix, and $ \{a\}_+ = \max(0, a) $. We further set $ ||\mathbf{X}||_F $ and $ ||\mathbf{X}||_* $ as the Frobenius norm and nuclear norm of a matrix $ \mathbf{X} $ respectively.

\section{Robust Matrix Regression}
We first introduce the RMR model to address the noisy label problem, with 
an ADMM algorithm for model training.
\subsection{Model}
For matrix regression, classical techniques need to reshape each matrix $ \mathbf{X}_i $ into a vector $ \mathbf{x}_i $, which will destroy its intrinsic structures, resulting in the loss of information. 
The R-GLM approach~\citep{ML2014_zhou_regularized} addresses this issue by enforcing the regression matrix $ \mathbf{W} $ to be low-rank representable with nuclear norm penalty.  
However, this method is based on the Gaussian loss, which may affect the robustness with existence of noisy labels.

To tackle this issue, an intuitive idea is to ignore noises within a small margin $ \{-\epsilon, \epsilon\} $ around each label for robust model fitting. 
Motivated by this idea, we propose our RMR, by introducing the hinge loss for model fitting, where the residual corresponding to each data $ \mathbf{X}_i $ is defined as follows
\begin{equation}
	\label{eqn:hinge_loss}
	h_i(\mathbf{W}, b) = (\lvert \text{tr}(\mathbf{W}^\top \mathbf{X}_i) + b - y_i \rvert - \epsilon)_+.
\end{equation} 

With the above formulation of residuals, when learning the regression model, our approach only takes residuals larger than $ \epsilon $ into account, thus, the labels contaminated by noise within a small margin is tolerable accordingly.
Similar residual modeling approach has also been used in the method of support vector regression~\citep{ML2004_smola_tutorial}.
However, this approach is proposed for vector data regression and cannot capture the latent structure within each data matrix. 
Differently, our method can capture such latent structure by incorporating the spectral elastic net penalty~\citep{ML2015_luo_support} into the regression matrix $ \mathbf{W} $, which can model the correlation of each data matrix effectively. 
And the corresponding optimization problem is defined as follows
\begin{equation}
	\label{eqn:optimization_label}
	\arg\underset{\mathbf{W},b}\min ~ ~  H(\mathbf{W},b)+ \tau || \mathbf{W} ||_*           \\
\end{equation}
where
\begin{align*}
	H(\mathbf{W},b) = ~ & \frac{1}{2}\text{tr}(\mathbf{W}^\top\mathbf{W})                              &           \\ 
	~  				  ~ & + C\sum_i^N \{-\epsilon-\text{tr}(\mathbf{W}^\top\mathbf{X}_i)-b+y_i\}_+     & \nonumber \\
	~ 			      ~ & + C\sum_i^N \{-\epsilon + \text{tr}(\mathbf{W}^\top\mathbf{X}_i)+b-y_i\}_+,  & \nonumber.
\end{align*}
with $\dfrac{1}{2}\text{tr}(\mathbf{W}^\top\mathbf{W}) + \tau || \mathbf{W} ||_*$  as the spectral elastic net penalty, we incorporate low-rank property into $ \mathbf{W} $ and consider the group effect of the eigenvalues, to capture the latent structures among data matrices~\citep{ML2015_luo_support}.

\subsection{Solver}
As the object function contains both hinge loss and nuclear norm, the Nesterov method used in R-GLM \citep{ML2014_zhou_regularized} is no longer available because the derivative of our loss function is not Lipschitz-continuous.
Nevertheless, since our model is convex with respect to both $\mathbf{W}$ and $b$, we here derive an efficient learning algorithm based on Alternating Direction Method of Multipliers~(ADMM)~\citep{ML2011_boyd_distributed} with the restart rule \citep{ML2014_goldstein_fast} to solve the optimization problem.  
The optimization problem defined in Eq.~\eqref{eqn:optimization_label} can be equivalently written as follows:

\begin{align}
	\label{eqn:subproblem1}
	\arg\underset{\mathbf{W},b,\mathbf{S}}\min   ~~ & ~ H(\mathbf{W},b)+G(\mathbf{S}) \\
	\mbox{s.t}. 								 ~~ & ~ \mathbf{S}-\mathbf{W} = 0 \nonumber
\end{align}
where $\mathbf{S}$ is an auxiliary variable and $G(\mathbf{S}) = \tau||\mathbf{S}||_*$. In this way, the original optimization has been split into two subproblems with respect to $ \{\mathbf{W}, b\} $ and $ \mathbf{S} $ respectively.
In this way, we can develop efficient ADMM method to solve Eq.~\eqref{eqn:subproblem1} by using the Augmented Lagrangian approach as follows:
\begin{align}
& & L(\mathbf{W},b,\mathbf{S},\mathbf{\Lambda}) =  & H(\mathbf{W},b) + G(\mathbf{S}) + \text{tr}[\mathbf{\Lambda}^\top(\mathbf{S}-\mathbf{W})] &  \nonumber \\
& & ~                 							   & + \frac{\rho}{2}||\mathbf{S}-\mathbf{W}||_F^2  &
\end{align}
where $\rho > 0$ is a hyper parameter.

\subsubsection{Optimization for Auxiliary Variable}

We first derive the optimization method for solving the auxiliary variable $\mathbf{S}$.
The first subproblem for solving $\mathbf{S}$ is
\begin{equation}
	\label{eqn:subproblem_1_1}  
	\arg\underset{\mathbf{S}}\min ~~ \tau || \rho \mathbf{S} ||_{*} + \dfrac{1}{2}||(\rho \mathbf{W} - \mathbf{\Lambda}) - \rho \mathbf{S}||_F^2.
\end{equation}
Then we can get $\mathbf{S}^{(k)}$ in the $k^{th}$ iteration by solving the problem in Eq.~\eqref{eqn:subproblem_1_1} and get the analytical solution as follows.
\begin{equation}
	\mathbf{S}^{(k)} = \dfrac{1}{\rho} \mathbf{U} \mathcal{D}_{\lambda}(\rho \mathbf{W}^{(k)} -  \mathbf{\Lambda}^{(k)})\mathbf{V}^\top,
\end{equation}
where $ \mathbf{U}\mathbf{\Sigma}\mathbf{V}^\top = \rho \mathbf{W}^{(k)} -  \mathbf{\Lambda}^{(k)}$ and $\mathcal{D}_{\lambda}(\mathbf{\Sigma})_{ii} = \max(\mathbf{\Sigma}_{ii} - \lambda, 0)$.

\subsubsection{Optimization for Regression Matrix}
To solve the regression matrix, we have
\begin{equation}
	\label{eqn:subproblem_1_2}  
	\arg\underset{\mathbf{W},b}\min ~ H(\mathbf{W},b) - \text{tr}(\mathbf{\Lambda^\top W}) + \dfrac{\rho}{2} ||\mathbf{W-S}||^2_F.
\end{equation}
And one solution of this optimization problem is
\begin{align}
\label{eqn:W}    
\mathbf{W^*} & = \frac{1}{1+\rho}(\sum_{i=1}^N (\alpha_i - \alpha_i^*) \mathbf{X}_i + \mathbf{\Lambda} + \rho \mathbf{S}),  &     \\ 
b^* & = \frac{1}{|\mathcal{I}^*|}\sum_{i \in \mathcal{I}^*}\{y_i- \emph{sign}(\beta_i^*)\cdot\epsilon - \emph{tr}[(\mathbf{W^*})^\top \mathbf{X}_i] \},                   & \nonumber
\end{align}
where $\mathcal{I}^* = \{i:0<\mathbf{\beta}_i^* <C\}$, $\mathbf{\beta}^* \in  \mathbb{R}^n$ and $\mathbf{\beta = \alpha - \alpha}^*$. 
$\mathbf{\alpha}$ and $\mathbf{\alpha^*}$ can be constructed from the solution of the following box constrained quadratic programming problem:
\begin{align}
\label{eqn:bqp}
\arg\underset{\mathbf{x}}\min ~~~& \frac{1}{2}\mathbf{x}^\top \mathbf{H} \mathbf{x} + \mathbf{c}^\top \mathbf{x}       	         	&           \\
\emph{s.t}                 ~~~&  \mathbf{0} \leq  \mathbf{x} \leq C \mathbf{1}_{2n}      	 & \nonumber \\
&  \sum_{i=1}^n (\mathbf{\alpha}_i - \mathbf{\alpha}_i^*) = 0.  & \nonumber
\end{align}
Here we have 
\begin{equation}
\mathbf{x} = \begin{bmatrix} \mathbf{\alpha} \\ ~~\mathbf{\alpha}^* \end{bmatrix} ,~~
\mathbf{c} = \begin{bmatrix} \mathbf{p}      \\ ~~\mathbf{p^*}      \end{bmatrix}, ~~
\mathbf{H} = \begin{bmatrix} \mathbf{K} & -\mathbf{K} \\ -\mathbf{K} & \mathbf{K}  \end{bmatrix}.  
\end{equation}
$\mathbf{K} = [K_{ij}] \in \mathbb{R}^{N \times N}$ and $q \in \mathbb{R}^N$ are independent of $\alpha$ with,
\begin{align*}
K_{ij} ~~~ = & ~~~ \dfrac{\emph{tr}(\mathbf{X}_i^\top \mathbf{X}_j)}{\rho + 1},  \\
p_i ~~~ = & ~~~ \epsilon - y_i + \dfrac{\emph{tr}[(\mathbf{\Lambda}+ \rho\mathbf{S})^\top\mathbf{X}_i ]}{\rho + 1}                                                        \\
p_i^* ~~~ = & ~~~ \epsilon + y_i - \dfrac{\emph{tr}[(\mathbf{\Lambda}+ \rho\mathbf{S})^\top\mathbf{X}_i] }{\rho + 1}     
	\end{align*}
In this way, we can get $\mathbf{W}^{(k)}$ and $b^{(k)}$ in each iteration with sequential minimization optimization algorithm \citep{ML1998_platt_sequential,ML2002_keerthi_convergence}. 

The algorithm is summarized in Algorithm~\ref{algo:sub1_admm}. And for its convergence, we have the following theorem.

\begin{mytheorem}
	Suppose the optimal solution of Problem (\ref{eqn:subproblem1}) is $(\tilde{\mathbf{W}}
	,\tilde{b},\tilde{\mathbf{S}})$. Then
	\begin{equation}
	\tilde{\mathbf{W}} = \tilde{\mathbf{S}} = \tilde{\mathbf{\Lambda}} + \sum (\tilde{\alpha}_i - \tilde{\alpha}_i^*) \mathbf{X}_i.
	\end{equation}
	
	\label{theo:sol}
\end{mytheorem}

Because the hinge loss and nuclear norm are weakly convex, the convergence property of Algorithm~\ref{algo:sub1_admm} can be proved immediately based on the result in \citep{ML2014_goldstein_fast, ML2012_he_on}. That is, we have

\begin{mytheorem}
	For any $\rho > 0$ and $\eta \in (0,1)$, the iteration sequence given by Algorithm~\ref{algo:sub1_admm} converges to the optimal solution of Problem~(\ref{eqn:subproblem1}).	
\end{mytheorem}

\begin{algorithm}[ht!]
	\caption{ADMM for Subproblem 1 in Eq.~\eqref{eqn:subproblem1}}
	\footnotesize
	\algrenewcommand\alglinenumber[1]{\scriptsize #1:}
	\begin{algorithmic}[1]
		\State Initialize $\mathbf{S}^{(-1)}=\widehat{\mathbf{S}}^{(0)}\in\mathbb{R}^{p \times q},~\mathbf{\Lambda}^{(-1)}=\widehat{\mathbf{\Lambda}}\in\mathbb{R}^{p\times q},~\rho>0,~t^{(1)}=1,~\eta \in (0,1).$
		\For{$k=0,1,2,3...$}
		\State {$(\mathbf{W}^{(k)},b^{(k)}) = \arg\underset{\mathbf{W},b}\min  H(\mathbf{W},b) - \text{tr}(\widehat{\mathbf{\Lambda}}^{(k)\top}\mathbf{W})+     \dfrac{\rho}{2}||\mathbf{W}-\widehat{\mathbf{S}}^{(k)}||_F^2$}
		\State {$ \mathbf{S}^{(k)}=\arg\underset{\mathbf{S}}\min~G(\mathbf{S})+ \text{tr}(\widehat{\mathbf{\Lambda}}^{(k)\top}\mathbf{S}) + \dfrac{\rho}{2}||\mathbf{W}^{(k)}-\mathbf{S}||_F^2 $}
		\State {$\mathbf{\Lambda}^{(k)} = \widehat{\mathbf{\Lambda}}^{(k)}-\rho(\mathbf{W}^{(k)}-\mathbf{S}^{(k)}) $}
		\State {$c^{(k)}=\rho^{-1}||\mathbf{\Lambda}^{(k)}-\widehat{\mathbf{\Lambda}}^{(k)}||_F^2 + \rho||\mathbf{S}^{(k)}-\widehat{\mathbf{S}}^{(k)}||_F^2 $}
		\If {$c^{(k)} < \eta c^{(k-1)} $}
		\State {$t^{(k+1)} = \dfrac{1+\sqrt{1+4^{(t)^2} }}{2}$}
		\State {$\widehat{\mathbf{S}}^{(k+1)} = \mathbf{S}^{(k)} + \dfrac{t^{(k)-1}}{t^{(k+1)}}(\mathbf{S}^{(k)} - \mathbf{S}^{(k)-1})  $}
		\State {$\widehat{\mathbf{\Lambda}}^{(k+1)} = \mathbf{\Lambda}^{(k)} + \dfrac{t^{(k)-1}}{t^{(k+1)}}(\mathbf{\Lambda}^{(k)} - \mathbf{\Lambda}^{(k)-1})  $}
		\Else
		\State {$t^{(k+1)} = 1$}
		\State {$\widehat{\mathbf{S}}^{(k+1)} = \mathbf{S}^{(k-1)}$}
		\State {$\widehat{\mathbf{\Lambda}}^{(k+1)} = \mathbf{\Lambda}^{(k-1)}$}
		\State {$c^{(k)} = \eta^{-1}c^{(k-1)}$}		
		\EndIf
		\EndFor 
	\end{algorithmic}
	\label{algo:sub1_admm}
\end{algorithm}

\subsection{Theoretical Analysis}
We further theoretically analyze the excess risk of our RMR model. Following the framework of \citep{ML2016_wimalawarne_theoretical}, we assume each entity of a data matrix follows standard Gaussian distribution. Then, the optimization problem of our RMR can be rewritten as 
\begin{align}
\arg&\underset{\mathbf{W},b}\min \sum_{i=1}^{N}l(\mathbf{W}, b, \mathcal{X}_i, y_i)	\\
&\text{s.t.} || \mathbf{W} ||_* \leq C_1, \qquad || \mathbf{W} ||_F \leq C_2, \nonumber
\end{align} 
where $ C_1 $ and $ C_2 $ are certain constants, and $ l(\mathbf{W}, b, \mathcal{X}_i, y_i) $ is the hinge loss. Based on the relation between $ \mathbf{W} $ and $ b $ in Eq.~(\eqref{eqn:W}), the loss function $ l(\mathbf{W}, b, \mathcal{X}_i, y_i) $ can be simplified as
\begin{align}
\hat{l}(\mathbf{W}, \tilde{\mathbf{X}}_i, y_i) 	& = \sum_i^N \{-\text{tr}(\mathbf{W}^\top \tilde{\mathbf{X}}_i) + c_3\}_+  \\
												& + \sum_i^N \{\text{tr}(\mathbf{W}^\top\tilde{\mathbf{X}}_i) + c_4\}_+, \nonumber
\end{align}
which is a $L$-\textit{Lipschitz continuous function}, and $ \tilde{\mathbf{X}}_i = \mathbf{X}_i - \dfrac{1}{N} \sum_{j = 1}^{N} \mathbf{X}_j $, with $\dfrac{1}{N} \sum_{j = 1}^{N} \mathbf{X}_j $ to be the empirical mean of data, which tends to be $ 0 $ when $ N $ is large. Thus, the $ \tilde{\mathbf{X}}_i $ can also be considered as standard Gaussian distributed.

Let $ R(\mathbf{W}) $ and $ \hat{R}(\mathbf{W}) $ be the empirical risk and expected risk respectively~\citep{ML2013_maurer_excess}. Also, we set $ \mathbf{W}^o $ be the optimal solution to
\begin{equation}
\mathbf{W}^o = \arg\underset{\mathbf{W}}\min R(\mathbf{W}), \qquad \text{s.t.} || \mathbf{W} ||_* \leq C_1, || \mathbf{W} ||_F \leq C_2,
\end{equation}
and $ \mathbf{\tilde{W}} $ be the optimal solution of 
\begin{equation}
\mathbf{\tilde{W}}  = \arg\underset{\mathbf{W}}\min \hat{R}(\mathbf{W}), \qquad \text{s.t.} || \mathbf{W} ||_* \leq C_1, || \mathbf{W} ||_F \leq C_2.
\end{equation}
Then, we provide the upper bound of the excess risk of RMR in the following theorem, with proof in supplementary. 
\begin{mytheorem}
	With probability at least $ 1 - \delta $, let $ r $ be the rank of $ \mathbf{W}^o $ ($ \mathbf{W}^o,\tilde{\mathbf{W}} \in \mathbb{R}^{p \times q} $), and the excess risk of RMR is bounded with
	\begin{align*}
	R(\tilde{\mathbf{W}}) - R(\mathbf{W}^o) &\leq\dfrac{2L \max \{C_1, \dfrac{1}{\sqrt{r}}C_2\}}{\sqrt{N}}\\
	&\cdot  (\sqrt{p} + \sqrt{q})  + \sqrt{\dfrac{\ln(1/\delta)}{2N}} 
	\end{align*}
\end{mytheorem}

\begin{myproof}
	To prove the above theorem, We can first reformulate the excess risk with respect to $ \mathbf{W}^o $ and $ \mathbf{\hat{W}} $ as follows
	\begin{equation}
	\begin{split}
	R(\mathbf{\tilde{W}}) - 	&R(\mathbf{W}^o) = [R(\mathbf{\tilde{W}}) - \hat{R}(\mathbf{\tilde{W}})]	\\
	&+ [\hat{R}(\mathbf{\tilde{W}}) - \hat{R}(\mathbf{W}^o)] + [\hat{R}(\mathbf{W}^o) - R(\mathbf{W}^o)]
	\end{split}
	\end{equation}
	Here, the second term is negative naturally. And following the \textit{Hoeffding's inequality}, the third one can be bounded as $ \sqrt{\ln(1/\delta) / 2 N} $, with probability $ 1 - \delta/2 $. 
	
	For the first term, it is easy to obtain that
	\begin{equation}
	R(\mathbf{\tilde{W}}) - \hat{R}(\mathbf{\tilde{W}}) \leq \underset{|| \mathbf{W} ||_{*} \leq C_1, || \mathbf{W} ||_{F} \leq C_2} \sup [R(\mathbf{W}) - \hat{R}(\mathbf{W})].
	\end{equation}
	Further using the \textit{McDiarmid's inequality}, we can simply obtain the \textit{Rademacher complexity} with probability $ 1 - \delta $, with
	\begin{equation}
	\mathcal{R} = \frac{2}{N} \mathbb{E}  \underset{|| \mathbf{W} ||_{*} \leq C_1, || \mathbf{W} ||_{F} \leq C_2} \sup \sum_{i=1}^{N}\sigma_i \hat{l}(\mathbf{W}, \tilde{\mathbf{X}}_i, y_i),
	\end{equation}
	where $ \sigma_i \in \{-1, 1\} $ represents the \textit{Rademacher variables}. Let $ \mathbf{\tilde{M}} = \sum_{i=1}^{N} \sigma_i \mathbf{\tilde{X}}_{i} $, we can obtain the upper bound of $ R(\mathbf{\tilde{W}}) - \hat{R}(\mathbf{\tilde{W}}) $ as follows
	\begin{equation}
	\begin{split}
	R(\mathbf{\tilde{W}}) - \hat{R}(\mathbf{\tilde{W}}) &\leq \mathcal{R}	\\
	&\leq\dfrac{2L}{N} \mathbb{E}  \underset{|| \mathbf{W} ||_{*} \leq C_1, || \mathbf{W} ||_{F} \leq C_2} \sup \sum_{i=1}^{N}\sigma_i\mbox{tr}(\mathbf{W}\mathbf{\tilde{X}}_i)	\\
	&= \dfrac{2L}{N} \mathbb{E}  \underset{|| \mathbf{W} ||_{*} \leq C_1, || \mathbf{W} ||_{F} \leq C_2} \sup \mbox{tr} (\mathbf{W} \mathbf{\tilde{M}}). 
	\end{split}
	\end{equation}
	Further applying the \textit{H\"{o}lder's inequality}, we have
	\begin{equation}
	\begin{split}
	R(\mathbf{\tilde{W}}) - \hat{R}(\mathbf{\tilde{W}}) & \leq \dfrac{2L}{N} \mathbb{E}  \underset{|| \mathbf{W} ||_{*} \leq C_1, || \mathbf{W} ||_{F} \leq C_2} \sup || \mathbf{W} ||_{*}|| \mathbf{\tilde{M}} ||_{*}^* 	\\
	& \leq \dfrac{2L}{N} \mathbb{E} \max(\underset{|| \mathbf{W} ||_{*} \leq C_1} \sup || \mathbf{W} ||_{*},\underset{|| \mathbf{W} ||_{F} \leq C_2} \sup || \mathbf{W} ||_{*} ) \\
	& \cdot || \mathbf{\tilde{M}} ||_{*}^* \\ 
	& = \dfrac{2L \max(C_1, \dfrac{1}{\sqrt{r}}C_2)}{N} \mathbb{E} || \mathbf{\tilde{M}} ||_{*}^* 
	\end{split}
	\end{equation}
	where $|| \mathbf{\tilde{M}}||_*^*$ is the dual norm of nuclear norm $|| \mathbf{\tilde{M}}||_*$ and
	\begin{equation}
	\begin{split}
	\sup 		~~  & || \mathbf{W} ||_{*} \\
	\mbox{s.t.} ~~  & || \mathbf{W} ||_{F} \leq C_2, 
	\end{split}
	\end{equation}
	is equivalent to
	\begin{equation}
	\begin{split}
	\sup 		~~  & \sum_i^r \sigma_i(\mathbf{W}) \\
	\mbox{s.t.} ~~  & \sum_i^r \sigma_i(\mathbf{W})^2 \leq C_2^2, 
	\end{split}
	\end{equation}
	whose optimal solution is $\dfrac{1}{\sqrt{r}}C_2$.
	
	Since $ \mathbf{\tilde{M}} $ is the sum of random variables, its entries should also be considered as Gaussian distributed, with variance $ \omega = N $. Thus, following \cite{ML2013_maurer_excess}, with the \textit{Gordan's theorem}, we have 
	\begin{equation}
	\begin{split}
	\mathbb{E} || \mathbf{\tilde{M}} ||_{*}^*  &\leq \sqrt{\omega} (\sqrt{p} + \sqrt{q})	\\
	&= \sqrt{N} (\sqrt{p} + \sqrt{q})
	\end{split}
	\end{equation}
	Combining all the above together, we can obtain the upper bound of the excess risk with probability at least $ 1 - \delta $ as follows
	\begin{equation}
	\begin{split}
	R(\tilde{\mathbf{W}}) - R(\mathbf{W}^o) &\leq\dfrac{2L \max \{C_1, \dfrac{1}{\sqrt{r}}C_2\}}{\sqrt{N}}\\
	&\cdot  (\sqrt{p} + \sqrt{q})  + \sqrt{\dfrac{\ln(1/\delta)}{2N}} 
	\end{split}
	\end{equation}
\end{myproof}
\section{Generalized Robust Matrix Regression}

\noindent

The RMR can tolerate label noises for matrix regression. However, it cannot handle noise or outliers on each predictor empirically.
To address this issue, we further introduce a generalize RMR~(G-RMR) which assumes each noisy data matrix can be decomposed as a latent clean signal plus outliers. The clean signals can be recovered from each noisy ones when learning the regression model. 

\subsection{Model}
As discussed in \citep{ML2011_candes_robust}, it is common for natural signals to contain correlation empirically. Thus, when stacking each vectorized latent clean data matrix, the resulting matrix should be low-rank representable. We also introduce the sparsity feature with the $ L_1 $ norm to the outliers for robust modeling.
With these settings, the optimization problem of G-RMR can be defined as follows
\begin{align}
	\label{eqn:optimization_entry}
	\arg\underset{\mathbf{W},b}\min ~ ~ & H(\mathbf{W},b) + \tau || \mathbf{W} ||_* + \gamma ||\mathbf{X} ||_* + \lambda ||\mathbf{E} ||_1   \\ 
	\text{s.t.}         ~  ~ & \mathbf{D} = \mathbf{X} + \mathbf{E} \nonumber. 
\end{align}
Here, $\mathbf{D}_i$ denotes the $ i $th noisy input matrix, and we assume that it can be decomposed as $\mathbf{D}_i = \mathbf{X}_i + \mathbf{E}_i$, with $\mathbf{X}_i  $ as the latent clean matrix signal, and $ \mathbf{E}_i$  as the outliers.
$\mathbf{D}$ is a matrix with the $i^{th}$ row as the vector form of $\mathbf{D}_i$,
$\mathbf{X}$ denotes the matrix with the $i^{th}$ row as the vector form of $\mathbf{X}_i$, which is assumed to a low-rank representable with a nuclear norm penalty; and $\mathbf{E}$ contains the outliers for each data matrix, with its $i^{th}$ row as the vector form of $\mathbf{E}_i$ and is encouraged to be sparse with the $ L_1 $ norm. It can be noticed that the decomposition form of each $ \mathbf{D}_i $ is the same as that in \textit{Robust Principal Component Analysis}, which is effective in data recovery. But our approach is different because we update the regression matrix and recover the clean signals simultaneously within one optimization problem, which can benefit both tasks empirically.

\subsection{Solver}

The optimization problem for G-RMR contain three additional non-smooth term.
Fortunately, the object function in Eq.~\eqref{eqn:optimization_entry} is still bi-convex, which means that it is convex with respect to $\mathbf{W}$ with $\mathbf{X}$ and $\mathbf{E}$ fixed, and convex with respect to $\mathbf{X}$ and $\mathbf{E}$ with $\mathbf{W}$ fixed.
Therefore, we derive an iterative ADMM method to solve this problem, where the optimization problem in Eq.~\eqref{eqn:optimization_entry} is divided into two subproblems and each subproblem can be solved by ADMM individually. 

The first subproblem is to solve $\mathbf{W}$ with $\mathbf{X}$ and $\mathbf{E}$ fixed, which is equivalent to solve the problem in Eq.~\eqref{eqn:optimization_label}. 
The second subproblem is to solve $\mathbf{X}$ and $\mathbf{E}$ with  $\mathbf{W}$ fixed. 
Different from the first subproblem, an $L_1$ norm is introduced to the object function.
And the subproblem can be written as
\begin{align}
	\label{eqn:subproblem2}
	\arg\underset{\mathbf{X},\mathbf{E}}\min ~~& H(\mathbf{W},b) + \gamma ||\mathbf{X}||_* + \lambda ||\mathbf{E}||_1, \\
	\mbox{s.t}.~~& \mathbf{D} = \mathbf{X} + \mathbf{E} \nonumber.
\end{align}

Since $H(\mathbf{W},b)$ contains hinge loss function and other two non-smooth terms, the optimization problem is hard to solve. 
Thus, we relax the hinge loss to squared one and develop another ADMM method to solve this subproblem in with Augmented Lagrangian approach as follows:
\begin{align}
	\label{eqn:subproblem2_lag}
	& & L(\mathbf{X},\mathbf{E},\mathbf{\Gamma}) =  & ~ C||\mathbf{X}\mathbf{w} + \mathbf{b} - \mathbf{y}||_2^2 + \gamma ||\mathbf{X}||_* + \lambda ||\mathbf{E}||_1 &  \nonumber \\
	& & ~										    &   + \text{tr}[\mathbf{\Gamma}^\top(\mathbf{D}-\mathbf{X}-\mathbf{E})]  								   &  \nonumber \\ 
	& & ~ 											&   + \frac{\mu}{2}||\mathbf{D}-\mathbf{X}-\mathbf{E}||_F^2~,  										   &  
\end{align}
where $\bf{w}$ is the coefficient variable in vector form, $\mathbf{b} = b \mathbf{e}_m$ $\bf{w}=\text{vec}(\mathbf{W})$ and $\mu$ is a hyper parameter.

We summarized our method to solve the problem in Eq.~\eqref{eqn:subproblem2_lag} in Algorithm~\ref{algo:sub2_admm}. The key steps include the computations of $\mathbf{X}^{(k)}$ and $\mathbf{E}^{(k)}$, where the derivation of both $\mathbf{X}^{(k)}$ and $\mathbf{E}^{(k)}$ are based on proximal gradient method.

We first solve $\mathbf{E}$ with $\mathbf{X}$ fixed, and the optimization problem is as follows,
\begin{align}
	\label{eqn:subproblem2_E}
	& & \arg\underset{\mathbf{E}}\min ~~    &  \lambda ||\mathbf{E}||_1 + \text{tr}[\mathbf{\Gamma}^\top(\mathbf{D}-\mathbf{X}-\mathbf{E})]  \\ 
	& & ~ 										 ~~   & + \frac{\mu}{2}||\mathbf{D}-\mathbf{X}-\mathbf{E}||_F^2~,\nonumber  &  
\end{align}
which can be further written as 
\begin{equation}
	 \arg\underset{\mathbf{E}}\min~~g_1(\mathbf{E}) +  h_1(\mathbf{E}),
\end{equation}
where $g_1(\mathbf{E}) = \dfrac{\mu}{2} ||\mathbf{D}-\mathbf{X}- \dfrac{1}{\mu}\mathbf{\Gamma}  - \mathbf{E}||_F^2$ and $ h_1(\mathbf{E}) = \lambda ||\mathbf{E}||_1 $.
Both of them are convex and their derivatives are Lipschitz continuous. Thus, we use the proximal gradient method to update $\mathbf{E}$ with,
\begin{equation}
\mathbf{E}^{(k)} = \text{Prox}_{t_k h}( \widehat{\mathbf{E}}^{(k)} - t_k \bigtriangledown g( \widehat{\mathbf{E}}^{(k)}))
\end{equation}
where $\text{Prox}_{t_k h_1}(\mathbf{X}) = \arg\underset{\mathbf{U}}\min (t_k h(\mathbf{U}) + \dfrac{1}{2}||\mathbf{U}-\mathbf{X}||_F^2)$ and $t_k$ is the size of a gradient step.

After solving $\mathbf{E}$,  we proceed to solve $\mathbf{X}$ with $\mathbf{E}$ fixed.
Similarly, we re-write the this subproblem as follows:
\begin{equation}
\arg\underset{\mathbf{E}}\min~~g_2(\mathbf{X}) +  h_2(\mathbf{X}),
\end{equation}
where $g_2(\mathbf{X}) = C||\mathbf{X}\mathbf{w} + \mathbf{b} - \mathbf{y}||_2^2 + \dfrac{\mu}{2}||\mathbf{D}-\mathbf{X}-\mathbf{E} - \dfrac{1}{\mu}\mathbf{\Gamma} ||_F^2$ and $h_2(\mathbf{X}) = \gamma ||\mathbf{X}||_*$.
Since both $g_2(\mathbf{X})$ and $h_2(\mathbf{X})$ are convex, $g_2(\mathbf{X})$ is smooth and $h_2(\mathbf{X})$ is non-smooth and their derivatives are  Lipschitz continuous, we can update $\mathbf{X}$ by
\begin{equation}
\mathbf{X}^{(k)} = \text{Prox}_{t_k h_2}( \widehat{\mathbf{X}}^{(k)} - t_k \bigtriangledown g( \widehat{\mathbf{X}}^{(k)})).
\end{equation}

\begin{algorithm}[ht!]
	\caption{ADMM for Subproblem 2 in Eq.~\eqref{eqn:subproblem2}}
	\footnotesize
	\algrenewcommand\alglinenumber[1]{\scriptsize #1:}
	\begin{algorithmic}[1]
		\State Initialize $\mathbf{X}^{(-1)}=\widehat{\mathbf{X}}^{(0)}\in\mathbb{R}^{n \times pq},~\mathbf{E}^{(-1)}=\widehat{\mathbf{E}}^{(0)}\in\mathbb{R}^{n \times pq},~\mathbf{\Gamma}^{(-1)}=\widehat{\mathbf{\Gamma}}\in\mathbb{R}^{n \times pq},~\rho>0,~t^{(0)}=1,~\eta \in (0,1), t_k \in [0,\frac{1}{L}], L~\text{is Lipschits constant}$.
		\For{$k=0,1,2,3...$}
		\State {$ \mathbf{E}^{(k)} = \text{Prox}_{t_k h_1}( \widehat{\mathbf{E}}^{(k)} - t_k \bigtriangledown g( \widehat{\mathbf{E}}^{(k)})) $}
		\State {$\mathbf{X}^{(k)} = \text{Prox}_{t_k h_2}( \widehat{\mathbf{X}}^{(k)} - t_k \bigtriangledown g( \widehat{\mathbf{X}}^{(k)}))$}
		\State {$\mathbf{\Gamma}^{(k)} = \widehat{\mathbf{\Gamma}}^{(k)}-\mu(\mathbf{D} - \mathbf{X}^{(k)}-\mathbf{E}^{(k)}) $}
		\State {$c^{(k)}=\mu^{-1}||\mathbf{\Gamma}^{(k)}-\widehat{\mathbf{\Gamma}}^{(k)}||_F^2 + \mu||\mathbf{X}^{(k)}-\widehat{\mathbf{X}}^{(k)}||_F^2 $}
		\If {$c^{(k)} < \eta c^{(k-1)} $}
		\State {$t^{(k+1)} = \dfrac{1+\sqrt{1+4^{(t)^2} }}{2}$}
		\State {$\widehat{\mathbf{X}}^{(k+1)} = \mathbf{X}^{(k)} + \dfrac{t^{(k)-1}}{t^{(k+1)}}(\mathbf{X}^{(k)} - \mathbf{X}^{(k)-1})  $}
		\State {$\widehat{\mathbf{\Gamma}}^{(k+1)} = \mathbf{\Gamma}^{(k)} + \dfrac{t^{(k)-1}}{t^{(k+1)}}(\mathbf{\Gamma}^{(k)} - \mathbf{\Gamma}^{(k)-1})  $}
		\Else
		\State {$t^{(k+1)} = 1$}
		\State {$\widehat{\mathbf{X}}^{(k+1)} = \mathbf{X}^{(k-1)}$}
		\State {$\widehat{\mathbf{\Gamma}}^{(k+1)} = \mathbf{\Gamma}^{(k-1)}$}
		\State {$c^{(k)} = \eta^{-1}c^{(k-1)}$}		
		\EndIf;
		\EndFor 
	\end{algorithmic}
	\label{algo:sub2_admm}
\end{algorithm}

We summarize our method for solving G-RMR in Algorithm~\ref{algo:all_iterative}, based on Algorithm~\ref{algo:sub1_admm} and Algorithm~\ref{algo:sub2_admm}, .

\begin{algorithm}[ht!]
	\caption{Solve Robust Matrix Regression}
	\footnotesize
	\algrenewcommand\alglinenumber[1]{\scriptsize #1:}
	\begin{algorithmic}[1]
		\State \textbf{Input:}~Input training data $(\mathbf{X}_1,y_1),...,(\mathbf{X}_n,y_n)$ and related parameters for the solver.
		\Repeat
			\State Update $(\mathbf{W},b)$ using Algorithm.~\ref{algo:sub1_admm};
			\State Update $\mathbf{X}$ and $\mathbf{E}$ using Algorithm.~\ref{algo:sub2_admm};
		\Until Convergence or maximum iteration number;
		\State \textbf{Output:}~The estimation of $(\mathbf{W},b)$.
	\end{algorithmic}
	\label{algo:all_iterative}
\end{algorithm}

\section{Experiments}
In this section, we conduct extensive experiments with comparative studies.
The proposed method is implemented by Matlab R2015a in a machine with four-core 3.7GHz CPU and 16GB memory. 
We investigate the performance of our RMR and G-RMR with comparison of two state-of-the-art methods: 1)~The classical Support Vector Regression~(SVR) \citep{ML2004_smola_tutorial}; 2)~Regularized Matrix Regression~(R-GLM) \citep{ML2014_zhou_regularized}.

To evaluate and compare the performance of these algorithms, we apply them on three empirical tasks. 
Firstly, we elaborate on the illustrative examples by examining different geometric and natural shapes on the regression matrix.
Secondly, we apply them on real-world finical time series data, where each sample can be represented as a matrix.
At last, we apply RMR and G-RMR on the application of human head pose estimation~\citep{ML2001_sherrah_fusion}. 

To evaluate the performance of each algorithm, we use the Relative Absolute Error (RAE) that measures error between true labels $\mathbf{y}$ and estimated labels $ \hat{\mathbf{y}}$ with $RAE_\mathbf{y} = ||\hat{\mathbf{y}}-\mathbf{y}||/{||\mathbf{y}||_2}$. 
For compared method, SVR, RMR and G-RMR, we fix the coefficients $C=1 \times 10^3$ and $\epsilon = 1 \times 10^{-2}$,.
And all the hyper-parameters are selected via cross validation.

\subsection{Shape Recovery}


We first conduct the illustrative examples by examining various signal shapes, where each of them represents a $64 \times 64$ regression matrix.
We use the regression matrix to generate each sample by the following equation:
\begin{equation}
y_i = \text{tr}(\mathbf{W}^\top\mathbf{X}_i) + b + \epsilon_i,
\end{equation}
where $\mathbf{W}$ is the regression matrix illustrated by a signal shape, $(\mathbf{X}_i,y_i)$ is a randomly generated sample, $b$ is a bias term and $\epsilon_i$ is the noise term on label $y_i$, which is sampled from Laplacian distribution (The probability distribution function is $P( x \lvert \mu,\sigma ) = \dfrac{1}{2 \sigma}\exp(-\dfrac{|x-\mu|}{\sigma}) $ ).

\begin{figure}[!t]
	\centering
	\includegraphics[width=1.0\columnwidth]{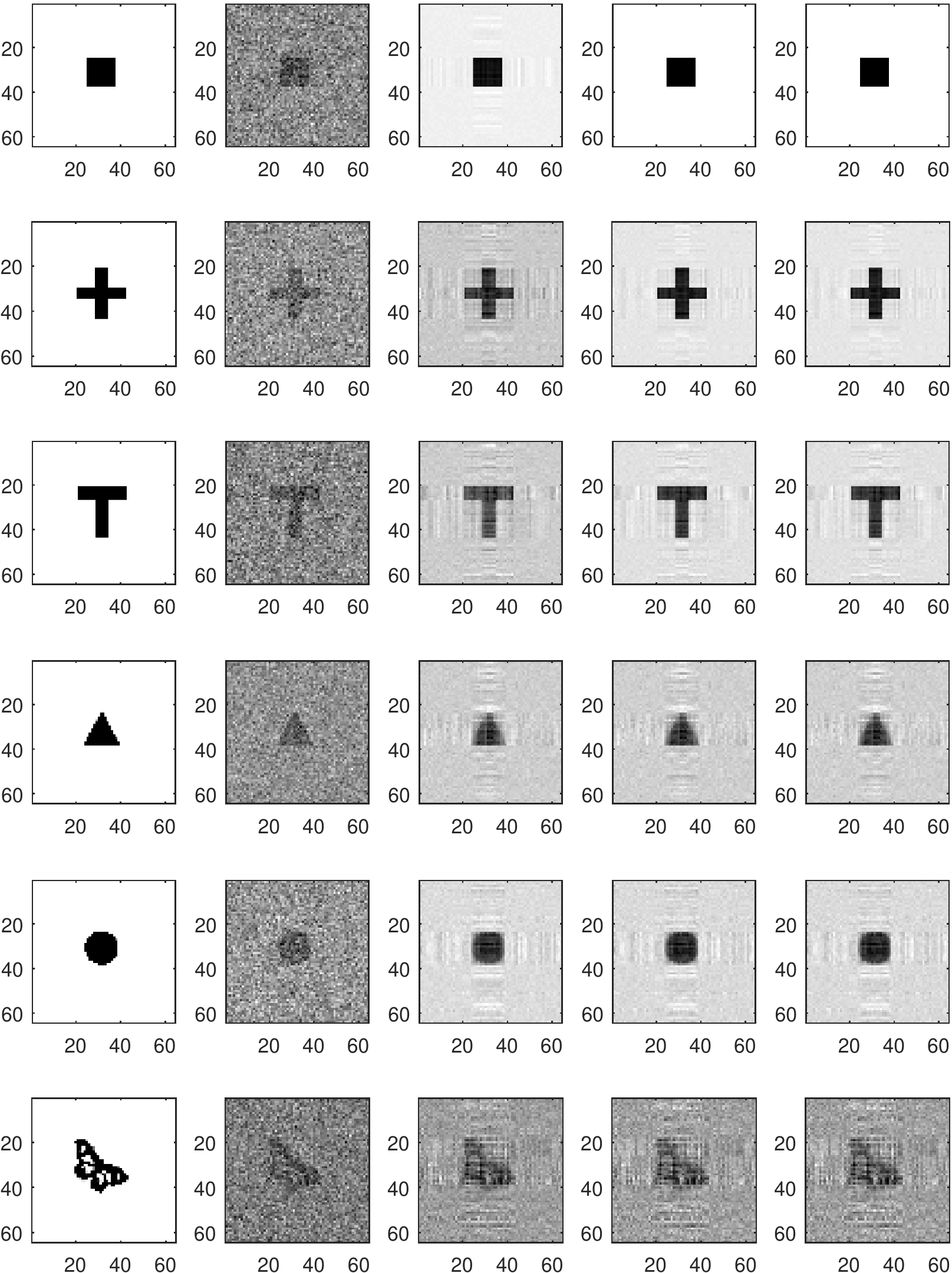}
	\caption{Comparison of our methods and other regression methods on the illustrative examples; Columns from left to right are true signal, SVR estimate, R-GLM estimate, RMR estimate and G-RMR estimate respectively.}
	\label{fig:shape}
\end{figure}

In the experiment, we randomly generate $1000$ samples for 10 rounds. 
In each round, half of the samples are used for model training and the rest are for testing.
Then we compute the mean and standard deviation of RAE error on classifier matrix $\mathbf{W}$ for each approach.
And detailed comparison of our RMR and G-RMR with other methods are shown in Table~\ref{table:shape}, where column  ``\textbf{shape}'' denotes the type of signals. 
The illustration of true signal shapes followed by the estimation results from the four methods can be found in the Fig.~\ref{fig:shape}.
\begin{table}[!th]
	\centering
	\vspace{-4ex}
	\caption{The RAE error on $\mathbf{W}$ for different approaches.}
	\vspace{1ex}
	\renewcommand{\arraystretch}{1.4}
	\resizebox{1\columnwidth}{!} {
		\begin{tabular}{l l l l l }
			\hline
			Shape               &\multicolumn{1}{l}{SVR}  &\multicolumn{1}{l}{R-GLM} &\multicolumn{1}{l}{RMR} &\multicolumn{1}{l}{G-RMR}         \\
			\thickhline                                                                              
			Square              &$0.9366\pm0.0047$    &$0.0658\pm0.0155$   &$\bf{0.0001\pm0.0001}$  &$0.0011\pm0.0004$       \\
			Cross               &$0.9377\pm0.0055$    &$0.4097\pm0.0619$   &$0.1678\pm0.0709$       &$\bf{0.1673\pm0.0706}$  \\
			T Shape             &$0.9373\pm0.0028$    &$0.4053\pm0.0365$   &$0.2059\pm0.0398$  		&$\bf{0.2045\pm0.0395}$  \\
			Triangle            &$0.9388\pm0.0039$    &$0.5375\pm0.0247$   &$0.5119\pm0.0220$  		&$\bf{0.5092\pm0.0216}$  \\
			Circle              &$0.9378\pm0.0048$    &$0.5375\pm0.0247$   &$0.3839\pm0.0179$  		&$\bf{0.3837\pm0.0178}$  \\
			ButterFly           &$0.9335\pm0.0036$    &$0.7319\pm0.0238$   &$0.7280\pm0.0245$  		&$\bf{0.7277\pm0.0245}$  \\
			\hline
		\end{tabular}
	}
	\label{table:shape}
\end{table}

\begin{figure}[h!]
    \centering
    \includegraphics[width=0.9\columnwidth]{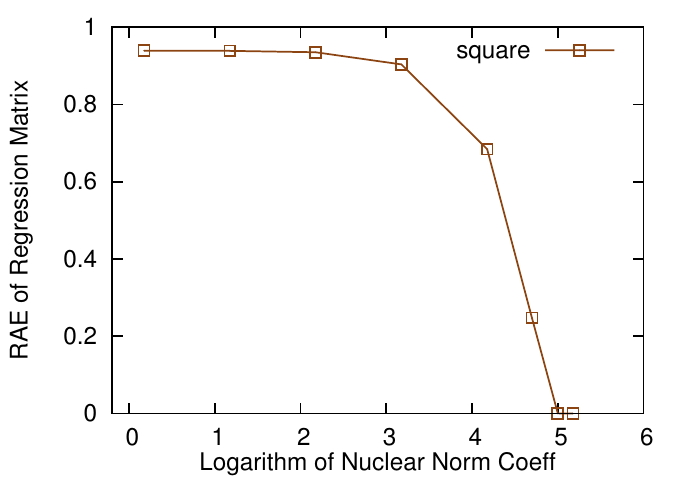}
    \caption{$RAE_{\mathbf{W}}$ along the solution path for nuclear norm regularization in RMR.}
    \label{fig:tau_path}
    \vspace{-2ex}
\end{figure}

It can be clearly seen from the illustrative examples that our method outperforms R-GLM in recovering lower rank signals, such as square, cross and T shape. 
Although they yield comparable results in recovering the high rank signal, it can be seen from Table~\ref{table:shape} that our approaches still outperform R-GLM in terms of the RAE error on $\mathbf{W}$ quantitatively.
It clearly shows that the hinge loss in our approaches for model fitting is more robust empirically.
Besides, it can be observed that our RMR and G-RMR methods substantially outperforms the traditional SVR method for all illustrative examples, because the SVR fails to capture the correlation among each data matrix.
Specifically, we further use the square shaped signal to display the RAE error along the solution path of the nuclear norm for matrix regression, which is shown in Fig.~\ref{fig:tau_path}. It can be seen that, when the weight is larger than $ 0 $, the performance is better, which shows the effectiveness of incorporating the nuclear norm as penalty.

\subsection{Financial Time Series Data Analysis}

We further evaluate our RMR on the financial time series data.
We use daily price data~(details can be found in the Table~\ref{table:ise_data}) from \citep{ML2014_akbilgic_a}, where the prices are converted to returns. 
In this data set, there are 536 daily returns from January 5, 2009 to Feburary 22, 2011. Particularly, the days are excluded when the Turkish stock exchange was closed. 

Intuitively, the value of an index in a certain date may be related to others as well as previous values.
Thus, it is natural to process data in matrix form instead of a vector to preserve the latent topological structure of data.
Besides, there exist many fluctuations in different indices due to the complicated stock market and many entries of the data are contaminated accordingly.
Therefore, it is imperative to tackle the above issues with the proposed methods.
\begin{table}[h!]
	\centering
	\caption{Abbreviations list for stock market indices}
	\vspace{1ex}
	\renewcommand{\arraystretch}{1.3}
	\resizebox{\columnwidth}{!}
	{
		\begin{tabular}{l l}
			\hline
			\tabincell{l}{Variable name}            &
			\tabincell{l}{Variable explanation}     \\
			\thickhline
			ISE100           & Istanbul stock exchange national 100 index     \\
			SP               & Standard \& poor’s 500 return index            \\ 
			DAX              & Stock market return index of Germany           \\
			FTSE             & Stock market return index of UK                \\ 
			NIK              & Stock market return index of Japan             \\ 
			BVSP             & Stock market return index of Brazil            \\ 
			EU               & MSCI European index                            \\ 
			EM               & MSCI emerging markets index                    \\ 
			\hline
		\end{tabular} 
	}
	\label{table:ise_data}
\end{table}

Besides using the RAE for evaluation, we further use 2 extra criterions to evaluate our results on the financial data set, i.e., percentage of correctly predicted (PCP) days, which can interpret our results in a simple and logical manner, 
and the Dollar 100 (D100) criterion, which gives us the theoretical future value of $\$100$ invested at the beginning of predicted term and traded accordingly.
We use the first $30\%$ days' data to train our model and predict the left $70\%$ days' index values for evaluation. 
\begin{table}[h!]
	\centering
	\vspace{-3ex}
	\caption{Results comparison on financial data}
	\vspace{1ex}
	\renewcommand{\arraystretch}{1.3}
	\resizebox{0.9\columnwidth}{!}
	{
		\begin{tabular}{l l l l l}
			\hline
			\tabincell{l}{Criterion}      &
			\tabincell{l}{SVR}            &
			\tabincell{l}{R-GLM}          &
			\tabincell{l}{RMR}         &
			\tabincell{l}{G-RMR}              \\
			\thickhline 
			PCF             & 55.43\%    & 54.62\%  &57.07\%    &\bf{58.42}\%  \\
			D100            & 163.5      & 122.2    &169.7      &\bf{195.2}    \\ 
			RAE             & 1.2729     &1.0160    &0.9914     &\bf{0.9868}   \\   
			\hline
		\end{tabular} 
	}
	\vspace{-2ex}
	\label{table:ise_result}
\end{table}

As we can see from Table~\ref{table:ise_result}, our proposed methods outperform other state-of-the-art methods in terms of all 3 different criterions.  
Our approaches achieve better results than previous R-GLM, because there exist many fluctuations in everyday's stock exchange rates due to the complicated situations, and allowing a range of error for the returns can be more robust than directly applying the squared error for model fitting.
Also, the SVR ignores the latent topological structure among different indices and historical data, and is beaten by our proposed methods.
It can be observed that our G-RMR achieves a significant improvement over the RMR. 
This is because financial data always contain serious noise problems~\citep{ML1998_magdon_financial} not only in the label but also in the data matrix entries. 
It shows that considering noise on predictors in G-RMR is also effective in certain real world situations.

\subsection{Head Pose Data Analysis}

\begin{figure}[ht!]
	\centering
	\includegraphics[width=0.8\columnwidth]{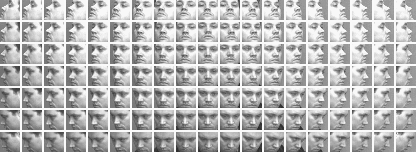}
	\caption{The example of the head pose estimation dataset.}
	\label{fig:head_pose}
\end{figure}
 
We further test the performance of our methods on the application of head pose estimation with dataset used in \citep{ML2001_sherrah_fusion} .
The dataset arises from a study to estimate the human pose via digital images captured from cameras, with 37 people in gray scale of $100 \times 100$ image size, which can be represented as data in matrix form naturally. 
Each person has $133$ facial images covering a view sphere of $\pm 90$ degrees in yaw and $\pm 30$ degrees in tilt at $10$ degrees increment. 
Several example images can be found in the Fig.~\ref{fig:head_pose}. 
As discussed before, it is difficult to measure the real angle of human head accurately, thus, it can be expected that the label of each image may contain several small errors, resulting in the noisy label problem.

For each person, we keep the degree in tilt fixed and use the degree in yaw angle $[0^\circ, 10^\circ,...170^\circ, 180^\circ]$ as our label. 
We then set tilting degree of the face to $90^\circ$, which denotes the frontal face.
Each image is cropped around the face and resized to $32\times32$.

\begin{table}[h!]
    \centering
    \caption{RAE error on the head angle of head pose dataset}
    \renewcommand{\arraystretch}{1.3}
    \vspace{2ex}
    \resizebox{0.9\columnwidth}{!}
    {
        \begin{tabular}{l l l l l l l}
            \hline
            \tabincell{l}{Trn\#}       &
            \tabincell{l}{SVR}         &
            \tabincell{l}{R-GLM}       &
            \tabincell{l}{RMR}         &  
            \tabincell{l}{G-RMR}       &    
            \tabincell{l}{G-RMR (C)}       \\
            \thickhline
            5    &0.3343    &0.2637  &\bf{0.2613} &\bf{0.2613} &0.2622   \\        
            10   &0.3996    &0.2355  &0.2195 	  &\bf{0.2194} &0.2292   \\        
            15   &0.3245    &0.2086  &0.2049 	  &\bf{0.2049} &0.2054   \\        
            20   &0.2918    &0.2042  &0.2029 	  &\bf{0.2029} &0.2040   \\        
            \hline
        \end{tabular} 
    }
    \label{table:head_pose}
\end{table}

The numerical performance of each algorithm is shown in Table.~\ref{table:head_pose},
where column ``\textbf{Trn\#}'' lists the corresponding number of training samples and column ``\textbf{G-RMR (C)}'' denotes that data used here is corrupted by adding white square blocks ($10\%$ samples are corrupted).
On one hand, as shown in Table~\ref{table:head_pose}, our methods outperform all competitive cones in terms of RAE error on the head angle, 
because our methods take both the correlation within each data and the noisy label issue into consideration, resulting in more robust estimation of the model. 
On the other hand, our G-RMR also achieves competitive result on corrupted data compared with the result on normal data, which shows the robustness of our model on noisy entries.

\section{Conclusions}
In this paper, we addressed the robust matrix regression issue with two novel methods proposed, i.e., RMR and G-RMR. For RMR, we introduced the hinge loss for model fitting, to enhance the robustness of matrix regression methods against the problem of noisy labels. An ADMM algorithm was further derived for model training. As an extension of RMR, the G-RMR was proposed to take noisy predictors into consideration by clean matrix signal recovery during model training procedure. We also conducted extensive empirical studies to evaluate the performance of RMR and G-RMR, and our methods achieve state-of-the-art performance. It shows that our approaches can address real-world problems effectively.

	\small                     
	\bibliographystyle{AAAI}
	\bibliography{../ref/top,../ref/ml,../ref/cv}


\end{document}